\documentclass[pdflatex,sn-mathphys]{sn-jnl}
\jyear{2023}
\theoremstyle{thmstyleone}%
\theoremstyle{thmstyletwo}%
\theoremstyle{thmstylethree}%
\usepackage{amsmath}
\usepackage{multirow}
\usepackage{breqn}
\usepackage{subcaption}

\usepackage[inkscapeformat=png]{svg}

\algnewcommand{\LeftComment}[1]{\Statex \(\triangleright\) #1}
\newcommand\norm[1]{\left\lVert#1\right\rVert}
\newcommand{\thicktilde}[1]{\mathbf{\tilde{\text{$#1$}}}}
\DeclareMathOperator*{\argmax}{arg\,max}

\DeclareMathOperator*{\tab}{\:\:\:\:\:\:\:}

\usepackage[symbol]{footmisc}

\begin{document}
\title[Similarity-based Knowledge Transfer for Cross-Domain RL]{Similarity-based Knowledge Transfer for Cross-Domain Reinforcement Learning}

\author*[1]{\fnm{Sergio A.} \sur{Serrano}}\email{sserrano@inaoep.mx}

\author[1,2]{\fnm{Jose} \sur{Martinez-Carranza}}\email{carranza@inaoep.mx}

\author[1]{\fnm{L. Enrique} \sur{Sucar}}\email{esucar@inaoep.mx}

\affil*[1]{\orgdiv{Computer Science Department}, \orgname{Instituto Nacional de Astrofísica, Óptica y Electrónica}, \orgaddress{\street{Luis Enrique Erro \#1}, \city{San Andrés Cholula}, \postcode{72840}, \state{Puebla}, \country{México}}}

\affil[2]{\orgdiv{Computer Science Department}, \orgname{University of Bristol}, \orgaddress{\street{Bristol BS8 1TL}, \city{Bristol}, \postcode{BS8 1TH}, \state{South West}, \country{England}}}

\abstract{Transferring knowledge in cross-domain reinforcement learning is a challenging setting in which learning is accelerated by reusing knowledge from a task with different observation and/or action space. However, it is often necessary to carefully select the source of knowledge for the receiving end to benefit from the transfer process. In this article, we study how to measure the similarity between cross-domain reinforcement learning tasks to select a source of knowledge that will improve the performance of the learning agent. We developed a semi-supervised alignment loss to match different spaces with a set of encoder-decoders, and use them to measure similarity and transfer policies across tasks. In comparison to prior works, our method does not require data to be aligned, paired or collected by expert policies. Experimental results, on a set of varied Mujoco control tasks, show the robustness of our method in effectively selecting and transferring knowledge, without the supervision of a tailored set of source tasks.\footnote[1]{This document is a preprint currently being under review for publication.}}



\maketitle

\section{Introduction}\label{sec:intro}
In recent years, Reinforcement Learning (RL) has proven to be a powerful tool capable of solving complex sequential decision-making problems with little supervision (\textit{e.g.} bipedal walking \cite{li2021reinforcement} and Atari video games \cite{mnih2013playing}), but at the cost of a large amount of data. For this reason, Transfer Learning (TL) methods have been used successfully to extend the applicability of RL to domains that cannot meet its data-cost requirements \cite{shao2018starcraft,hua2021learning}, usually reusing knowledge (\textit{e.g.} value functions \cite{serrano2021,lecarpentier2021lipschitz}, state transitions \cite{rusu2015policy}, policies \cite{abel18b}) in a more complex but related task (\textit{e.g.} learning to walk in a hexapod robot with help of a quadruped trained agent \cite{wan2020mutual}).

To transfer knowledge in a cross-domain setting (\textit{e.g.} robots with different morphology), TL methods face an additional challenge (compared to the intra-domain scenario) in finding an equivalence between observations from tasks with different states and/or actions (see Fig. \ref{fig:task-taxonomy}). A common practice is to assume the source and target task share a high level structure that conditions in some form their dynamics. Thus, most cross-domain TL works focus on aligning the information transferred, whether a cross-domain mapping is learned \cite{taylor2008autonomous,ammar2013automatically,zhang2020learning,you2022cross}, or a set of model-dependent parameters is reused \cite{rusu2016progressive}.

In either case, this assumption limits the range of applications to problems in which an expert \textit{a priori} selects the source of knowledge. For instance, in an RL database \cite{ramos2021rlds} to which users can upload small data sets sampled from real robots, it would be useful to transfer policies trained to solve similar tasks, and avoid the wear and tear caused by a full training process \cite{kober2013reinforcement}. However, as the rate at which data sets are uploaded increases, it becomes impossible to manually analyze and identify shared structures or task redundancy.

On the other hand, given that having common features across tasks is a cornerstone in TL, some works measure the inter-task similarity  and use it as a heuristic to select the best (\textit{i.e.} most similar) source of knowledge. However, most similarity based works address the intra-domain setting \cite{ammar2014automated,narayan2019effects,carroll2005task,dizicheh2022effective}, while those that operate in the cross-domain scenario require performing RL training in the target task \cite{serrano2021,carroll2005task}, which renders them not helpful as they can only be used when the target task has been learned.

\begin{figure}
    \centering
    \includegraphics[width=\textwidth]{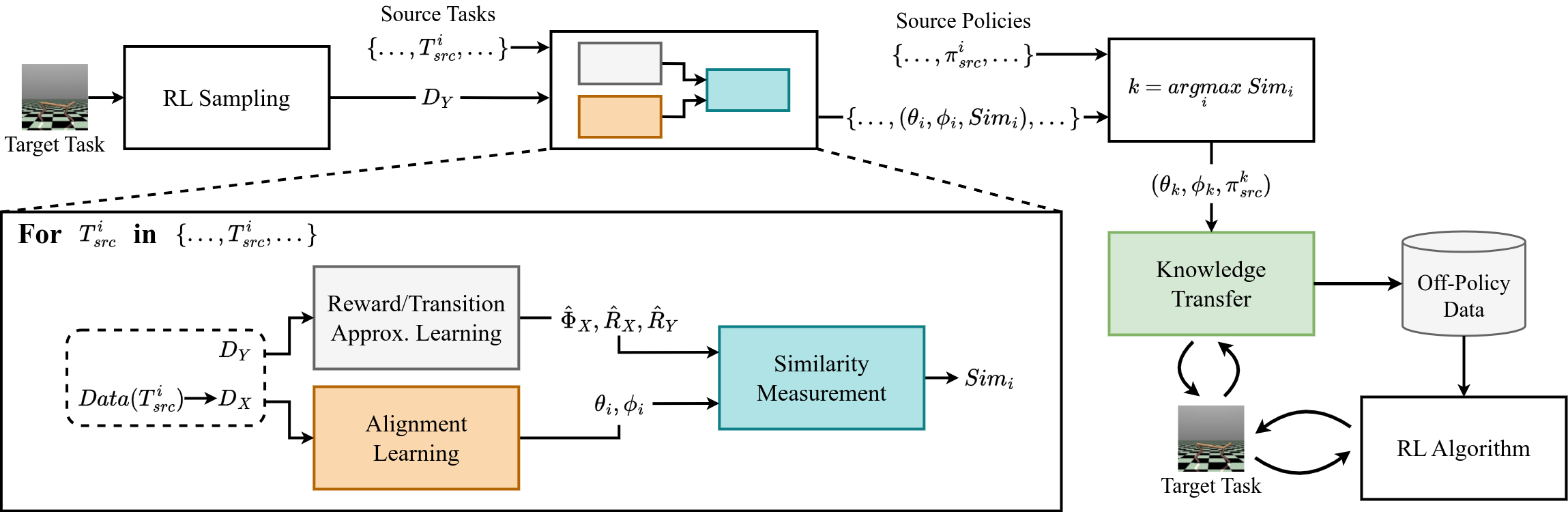}
    \caption{Overview of \textit{SimKnoT}. Given a target task and set of source tasks (Section \ref{sec:problem-setting}), a reward-based alignment between the target and each source state-action space is learned (Section \ref{sec:alignment}). Then, the alignment models (along with a set of approximated reward/transition models) are used to measure the target-source similarity based on their transition and reward dynamics (Section \ref{sec:task-simi}). Finally, actions are drawn from the most similar source task's policy and transferred to the target agent (Section \ref{sec:knowledge-transfer}), which generates an off-policy data set that an RL algorithm can use to speed up the learning process.}
    \label{fig:method}
\end{figure}

To bridge the gap between cross-domain TL and intra-domain similarity works, we propose a similarity-based knowledge transfer algorithm called \textit{SimKnoT} which not only is capable of transferring knowledge between tasks with different state-action spaces (after learning a cross-domain mapping), but it also selects the best source of knowledge (based on the similarity of the reward and transition dynamics), from a varied set of potential candidates. Thus, we set out to make the following contributions (see Fig. \ref{fig:method} for an overview):
\begin{enumerate}
    \item A reward-based semi-supervised loss function (\textit{ReBA}) that allows training a pair of encoder-decoder network pairs to align the state-action spaces of tasks with different representations.

    \item A similarity function that uses the \textit{ReBA}-trained models to rank a set of RL learned tasks, based on how similar their reward and transition functions are to the target task dynamics.

    \item A knowledge transfer algorithm (\textit{SimKnoT}) that uses the \textit{ReBA}-trained models to map states to the source domain, and actions to the target task, to reduce the training data cost by transferring source-optimal actions.
\end{enumerate}

The rest of the article is structured as follows, Section \ref{sec:background} introduces the main concepts of Reinforcement Learning, Transfer Learning and a component-wise taxonomy of inter-task similarity. Section \ref{sec:related-work} summarizes a review of the most related research, as well as a qualitative comparison to our work. Section \ref{sec:problem-setting} describes the problem setting and scope of our research, while Sections \ref{sec:alignment} and \ref{sec:cross-domain} introduce in detail the proposed alignment learning method and the similarity-based knowledge transfer algorithm. In Section \ref{sec:exp} the experimental setting and results are shown. Finally, Section \ref{sec:conclusions} presents our final remarks and future work.

\section{Background}\label{sec:background}
\subsection{Reinforcement Learning}\label{sec:rl}
Reinforcement Learning is the subarea of machine learning concerned with solving sequential decision-making problems. In RL, problems are modeled as Markov Decision Processes (MDP) that consist of a tuple $\langle S,A,\Phi,R,\gamma \rangle$ wherein the sets of variables $S_1,...,S_N \in \mathbb{R}$ and $A_1,...,A_M \in \mathbb{R}$ describe the state space $S=\{S_1 \times ... \times S_N\}$ and action space $A=\{A_1 \times ... \times A_M\}$, which represent the states the system can be at and the actions the agent can perform to change the system state, respectively. The transition function $\Phi : S \times A \times S \rightarrow [0,1]$ specifies the probability that the system will change from one state to another after performing a given action. The reward function $R : S \times A \rightarrow \mathbb{R}$ returns the immediate reward that the agent will receive after executing an action at a certain state, while the discount factor $\gamma \in [0,1)$ specifies the present value of future rewards \cite{puterman2014markov}.

When the transition and reward functions are not known, the agent must interact with its environment by executing actions and observing the immediate reward and next state. In model-free learning, observations are used to update a policy (\textit{e.g.} Q-learning \cite{watkins1992q}, SARSA \cite{sutton2018reinforcement}), where the goal is to learn an optimal policy $\pi^\ast : S \rightarrow A$ (Eq. \ref{eq:opt-policy}) that maximizes the state-value function, defined by Bellman's equation (Eq. \ref{eq:bellman}), which recursively defines the expected discounted reward based on the immediate reward $R(s,a)$ and next-state probability $\Phi(s' \mid s,a)$.

\begin{equation}\label{eq:bellman}
    V^{\pi}(s) = max_a \bigg\{ R(s,a) + \gamma \sum_{s' \in S} \Phi(s' \mid s,a) V^{\pi}(s') \bigg\}
\end{equation}

\begin{equation}\label{eq:opt-policy}
    \pi^{\ast}(s) = argmax_a \bigg\{ R(s,a) + \gamma \sum_{s' \in S} \Phi(s' \mid s,a) V^{\pi^\ast}(s') \bigg\}
\end{equation}

\subsection{Transfer Learning}\label{sec:tl}
Transfer learning focuses on studying methods that reuse and generalize knowledge from a previous problem (\textit{source task}) to learn better or faster in a new one (\textit{target task}). In RL, the objective of transferring knowledge is to reduce the data required to learn to solve a task (\textit{i.e.} sample complexity) and/or raise the agent's performance (\textit{i.e.} positive transfer). Conversely, the phenomenon of sample complexity increasing (or the agent's performance getting worse) due to the knowledge transfer process is known as negative transfer.

The effectiveness of TL methods (in RL) can be measured with multiple evaluation metrics: the \textit{jumpstart} is the difference in initial performance, the \textit{asymptotic performance} describes the performance at the end of the learning process, the \textit{total reward} is the total reward accumulated by the learning agent, the \textit{transfer ratio} represents the relative difference in performance (caused by transferring knowledge) with respect to the RL agent, and the \textit{time to threshold} is the difference in the sample complexity required to achieve a given performance \cite{taylor2009transfer}.

\subsection{Inter-Task Similarity}\label{sec:its}
Reinforcement learning tasks that are modeled as MDPs can be categorized according to the components they have in common. Intra-domain pairs (\textit{i.e.} tasks that share the same state-action space) can be classified, depending on their dynamics, as identical (their transition and reward functions are the same) and similar (presence of differences in the transition and/or reward function).

\begin{figure}
    \centering
    \includegraphics[width=\textwidth]{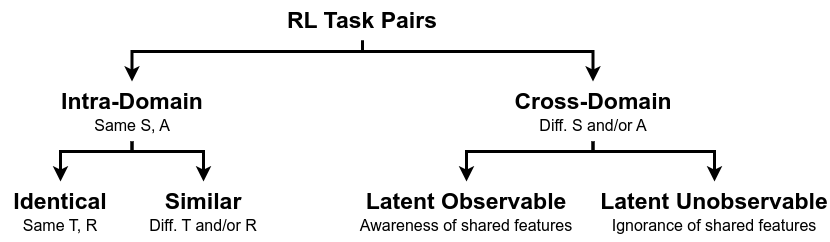}
    \caption{Taxonomy of RL settings based on the component-wise similarity between a pair of tasks. In the case of cross-domain settings, they are categorized based on the knowledge the human/expert has about inter-task common features that are not expressed as common state-action variables.}
    \label{fig:task-taxonomy}
\end{figure}

Alternatively, cross-domain pairs (\textit{i.e.} tasks that are defined over different representations) can have a common latent structure that facilitates transferring knowledge. However, considering the difficulty of determining the presence of such structure before the transfer experiment, we classify cross-domain pairs as latent observable and latent unobservable, depending on whether a successful transfer is expected given domain-specific knowledge (\textit{e.g.} a pair of robots with similar morphology but different number of limbs), or there are no reasons to assume such result will occur (see Fig. \ref{fig:task-taxonomy}).

Given that inter-task similarity is one of the main reasons to expect good results when knowledge transfer is performed, attempting to transfer knowledge in a \textit{cross-domain latent unobservable} setting may seem unproductive. However, similarly to how not every pair of \textit{similar} MDPs guarantees a positive transfer, it is reasonable to assume that some tasks with a mismatch in their representation have common features that can be exploited by a TL method (despite being beyond the expert's knowledge/scope), such as cart pole and mountain car \cite{ammar2014automated}, or robots with different morphologies \cite{raychaudhuri2021cross}.

\section{Related Work}\label{sec:related-work}
\paragraph{Similarity Measures}
In order to learn better/faster reusing knowledge, it is necessary that the source and target tasks have common attributes. Similarity measure works study ways in which similarity definitions can help identify good sources of knowledge to transfer from. In \cite{garcia2022taxonomy}, similarity measures that compare elements of an MDP are called model-based. For instance, in scenarios where only the transition \cite{ammar2014automated,narayan2019effects} or reward function change \cite{carroll2005task,dizicheh2022effective} (not both), comparing a slice of the environment dynamics can be enough to benefit from transferring state-transition tuples and policies. Some works define equivalence classes based on the MDPs transition and reward models, such as Bisimulation \cite{ferns2004metrics,ferns2012metrics,song2016measuring} and MDP homomorphisms \cite{sorg2009transfer,ravindran2004algebraic}, whereas in \cite{wang2019measuring} the dynamics are represented as an MDP-based graph that is compared with others using structural similarity methods.

On the other hand, performance-based works define similarity based on policies and their performance in both the source and target task \cite{garcia2022taxonomy}\footnote{For a more detailed and complete analysis of similarity measures for MDPS, a revision of the survey presented by \cite{garcia2022taxonomy} is strongly suggested.}. By computing the policy overlap \cite{carroll2005task} and value function differences \cite{carroll2005task,serrano2021} it is possible to compare tasks with different state-action spaces. Moreover, \cite{lecarpentier2021lipschitz} shows that (for the intra-domain setting) the optimal action-value function is Lipschitz continuous, which allows establishing upper bounds for the difference between two policies and is used as a source-selection heuristic. However, while most of the reviewed similarity measures focus on the intra-domain scenario, the few works that are cross-domain compatible \cite{carroll2005task,serrano2021} are also restricted to discrete spaces and require (partially) learning the target policy, which is too late for knowledge-transfer purposes.

\paragraph{Representation Learning for Knowledge Transfer}
Learning representations, that capture the common features of a set of tasks, can help to learn multiple tasks faster as knowledge is transferred across tasks through the new representation \cite{ammar2014online,zhao2017tensor}, or to better select the most similar source task from which knowledge will be transferred \cite{zhou2020latent}. By formalizing the latent-representation learning problem for a family of MDPs (as Hidden Parameters MDPs \cite{konidaris2014hidden}) the latent representation approach has been used to transfer value functions \cite{doshi2016hidden}, whereas generalization over unseen tasks has been achieved by embeddings learned based on a bisimulation metric \cite{agarwal2021contrastive,bertran2022efficient} and the characteristic function of reward sequence distributions \cite{yang2022learning}. Similarly, the Multi-view RL setting \cite{li2019multi} (observation-extended version of the Partially Observable MDPs \cite{kaelbling1998planning,serrano2021knowledge}) builds a latent space to align observations of a single phenomenon, from multiple points of view.

To transfer knowledge in the cross-domain setting, latent spaces have been learned to align the transition dynamics and transfer knowledge in the form of trajectories of expert demonstrations \cite{raychaudhuri2021cross}, policies without any further tuning \cite{zhang2021learning}, weighted influence from preactivation layers of source policies \cite{wan2020mutual,heng2022crossdomain}, or auxiliary rewards based on the source and target transition divergence \cite{gupta2017learning}. However, special data recollection requirements may narrow the applicability of some methods, whether it is pairing and aligning transitions \cite{gupta2017learning}, or sampling demonstrations with expert policies for the same task in two domains \cite{raychaudhuri2021cross}. Moreover, while \cite{heng2022crossdomain} enables transferring knowledge from multiple sources, it assumes the set of source tasks and target task share a high-level structure that conditions their dynamics (see Table \ref{tab:rw-comparison} for a comparison of latent-based methods).

Inspired by the success of similarity measures (in the intra-domain setting) as source-selecting heuristics, and latent-based TL methods (in the latent-observable setting), our knowledge transfer method borrows ideas from both areas to learn a reward-based latent space (with on-policy RL data) to measure similarity with a varied set of source tasks, dispensing the shared high-level structure assumption. Thus, the similarity-based knowledge transfer provides the benefits of TL, without increasing the little supervision required by RL methods.

\begin{table}[t]
\caption{Comparison of representation learning knowledge transfer works. The second column shows the task differences that are allowed, which include: state (S) and action (A) spaces, reward (R) and transition (T) functions, and observation space-function (O), while multiple working modes are separated by a forward slash. The last two columns show how many source tasks are allowed, and whether the method assumes the source and target tasks share a high-level structure (SHLS) that conditions their reward and transition dynamics. The last column is not applicable (N/A) to methods that assume a shared state-action.}
\label{tab:rw-comparison}
\centering
\begin{tabular}{ccccc}
\hline
Work                            & Differences  & Sources   & Assume SHLS\\ \hline
\cite{doshi2016hidden}          & R, T         & multi     & N/A       \\
\cite{agarwal2021contrastive}   & R, T         & multi     & N/A       \\
\cite{bertran2022efficient}     & R, T         & multi     & N/A       \\
\cite{zhou2020latent}           & R, T         & multi     & N/A       \\
\cite{ammar2014online}          & R, T         & multi     & N/A       \\
\cite{zhao2017tensor}           & R, T         & multi     & N/A       \\
\cite{yang2022learning}         & R, T         & multi     & N/A       \\
\cite{li2019multi}              & O            & multi     & N/A       \\
\cite{wan2020mutual}            & S, A         & single    & yes       \\
\cite{chen2019learning}         & S, A         & single    & yes       \\
\cite{gupta2017learning}        & S, A         & single    & yes       \\
\cite{zhang2021learning}        & S, A / O, T  & single    & yes       \\
\cite{raychaudhuri2021cross}    & S, A         & single    & yes       \\
\cite{heng2022crossdomain}      & S, A         & multi     & yes       \\
Ours                            & S, A         & multi     & no        \\ \hline
\end{tabular}
\end{table}

\section{Problem Setting}\label{sec:problem-setting}
Let $T_{src}^i=\langle S_{src}^i,A_{src}^i,\pi_{src}^i,D_{src}^i \rangle$ be a source task, where $S_{src}^i\times A_{src}^i$ is the state-action space, $\pi_{src}^i$ a trained policy and $D_{src}^i \subseteq S_{src}^i \times A_{src}^i \times \mathbb{R} \times S_{src}^i$ the set of \textit{state-action-reward-next state} tuples sampled during the training process of $\pi_{src}^i$. Also, let $T_{tgt}=\langle S_{tgt},A_{tgt},E_{tgt} \rangle$ be a target task, where $S_{tgt}\times A_{tgt}$ is also the task state-action space and $E_{tgt} : A_{tgt} \rightarrow \mathbb{R} \times S_{tgt} \times \{0,1\}$ is the environment function, which stores the current state internally, takes an action as input and returns the immediate reward, the next state and a binary flag indicating if the returned state is an initial state (\textit{i.e.} the episode has restarted).

Considering a set of $N$ source tasks $T_{src}=\{ T_{src}^1,...,T_{src}^N \}$ and a target task $T_{tgt}$ (where the task-pair $(T_{src}^i, T_{tgt})$ may belong to any of the four inter-task similarity categories presented in Sec. \ref{sec:its}), the problem consists of learning the target task $T_{tgt}$ as fast as possible, with the option to transfer knowledge from any subset of $T_{src}$, without incurring in negative transfer (see Sec. \ref{sec:tl}).

\subsection{Scope}\label{sec:scope}
The research problem is bounded by the following restrictions:
\begin{itemize}
    \item \textbf{Episodic Tasks}: tasks have a maximum number of steps that can be taken before the episode ends and the environment state resets to an initial state.
    \item \textbf{Fully Observable State}: observations completely describe the environment's state, thus, $observation=state$.
    \item \textbf{Stationary Dynamics}: reward and transition models do not change over time.
    \item \textbf{Continuous Spaces}: state and action spaces are assumed to consist of the cross product of a set of continuous variables.
    \item\textbf{Bounded Actions and Observations}: each action, state, and reward variable is bounded, thus their lower and upper limits are known \textit{a priori} for all tasks.
    \item\textbf{Dense/Rich Rewards}: tasks with reward models that are a dense function of their state-action space.
\end{itemize}

\section{Cross-Domain Alignment}\label{sec:alignment}
In order to transfer knowledge across tasks with different representations, an equivalence must be found between the source and target states/actions, such that enables transferring source knowledge (\textit{e.g.} policies, value functions, data \cite{taylor2009transfer}) to the target learner and improves its learning efficiency. Matching a pair of tasks, with no prior knowledge of how they might be related to each other (\textit{i.e.} cross-domain latent unobservable scenario, see Fig. \ref{fig:task-taxonomy}), requires finding and using common features to align a pair of domains. In the following sections, a learning objective for the domain alignment problem is presented (Section \ref{sec:rebal}), as well as the learning procedure to train a set of neural networks to perform the cross-domain alignment (Section \ref{sec:learn-rebal}).

\subsection{Reward-Based Alignment Loss}\label{sec:rebal}
\begin{figure}
    \centering
    \includegraphics[width=\textwidth]{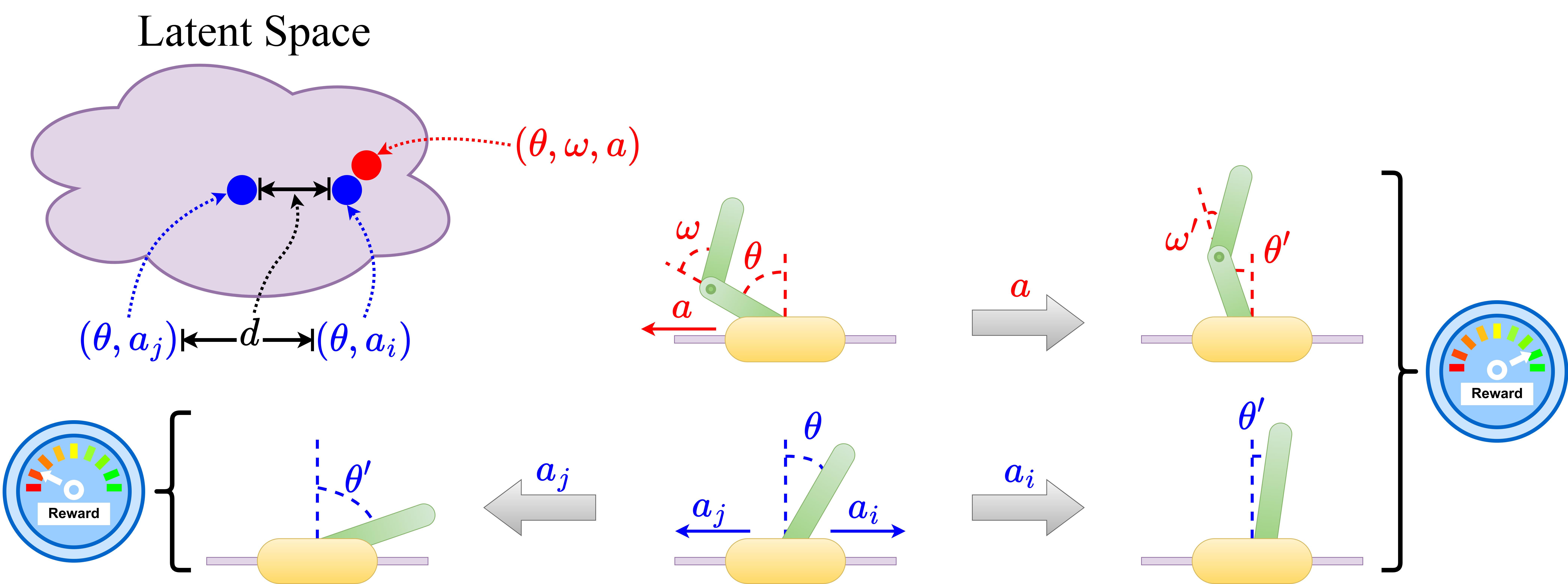}
    \caption{The reward-based aligned latent space joins state-action pairs from different domains that produce similar immediate rewards (Eq. \ref{eq:alignment}), while the distance among state-action pairs from the same task is preserved (Eq. \ref{eq:geo-pre}). For instance, in the single and double inverted pendulum tasks, cart velocities that are able to erect the poles (\textit{e.g.} $a,a_i$) yield high-reward transitions, and are mapped close to each other. The relative distance between low and high reward transitions (\textit{e.g.} $(\theta,a_j)$ and $(\theta,a_i)$) is maintained in the latent space.}
    \label{fig:intuition}
\end{figure}

In RL, the immediate reward signal provides a strict ordering of every state-action pair, and has the same interpretation across all RL tasks: \textit{the higher the reward, the more desirable a state-action pair}. When no correspondences are known to exist among RL tasks, the normalized immediate reward offers a reference that can be exploited to align the state-action spaces, and use the alignment to transfer knowledge (see Fig. \ref{fig:intuition}).

In similar fashion to \cite{yang2022learning}, we propose to use the reward signal as the matching criterion to guide the alignment of two observation-action spaces (in our case to address the cross-domain setting). Moreover, our proposal is inspired by the work of \cite{wang2009manifold} in which, through the minimization of a cost function (Eq. \ref{eq:uma}), two manifolds are aligned by matching similar local neighborhood patterns across manifolds in a new latent space, as well as maintaining the local neighborhoods existent in the original spaces.

\begin{align}
    \begin{split}\label{eq:uma}
    C(\alpha,\beta) = \mu \sum_{i,j}(\alpha^Tx_i-\beta^Ty_j)^2W^{i,j}
    & + 0.5 \sum_{i,j}(\alpha^Tx_i-\alpha^Tx_j)^2W^{i,j}_x \\
    & + 0.5 \sum_{i,j}(\beta^Ty_i-\beta^Ty_j)^2W^{i,j}_y
    \end{split}
\end{align}

Besides maintaining a set of linear restrictions, linear mappings $\alpha \in M_{p \times m}(\mathbb{R}), \beta \in M_{q \times m}(\mathbb{R})$ minimize their cost $C(\alpha,\beta)$ by mapping inter-domain ($x_i,y_j$) and intra-domain ($x_i,x_j$ and $y_i,y_j$) pairs closer in the latent space, the more similar they are. In Eq. \ref{eq:uma}, the first term penalizes mapping far apart elements (from different domains) that have a high similarity, determined by $W^{i,j}$. The second and third term penalize the same behavior among points from domains $X$ and $Y$, based on the intra-domain similarity scores $W^{i,j}_x,W^{i,j}_y$, respectively. A constant coefficient $\mu$ weights the importance of the inter-domain alignment against the intra-domain counterpart.

To train a pair of non-linear function approximators (\textit{e.g.} neural network) to learn a cross-domain alignment by the minimization of Eq. \ref{eq:uma}, there is a risk of learning a trivial solution that maps every point to a single location of the latent space (as their non-linear expressiveness does not restrict them). To overcome this hurdle, a reward-based learning loss is proposed, such that promotes the alignment of state-actions with similar rewards, as well as maintaining the local neighborhoods present in the original spaces.

Let $\theta_X : \mathbb{R}^p \rightarrow \mathbb{R}^m$ and $\theta_Y : \mathbb{R}^q \rightarrow \mathbb{R}^m$ be a pair of trainable functions, and $D_x,D_y$ a pair of normalized training sets defined as follows:
\begin{align*}
    &D_x=\{(x,r^x)|x \in \mathbb{R}^p, \: r^x \in [0,1], \: x[i] \in [0,1] \text{ for } i=1,...,p \} \\
    &D_y=\{(y,r^y)|y \in \mathbb{R}^q, \: r^y \in [0,1], \: y[i] \in [0,1] \text{ for } i=1,...,q \}
\end{align*}
then, the alignment loss is computed by Eq. \ref{eq:alignment}:
\begin{align}
    \begin{split}\label{eq:alignment}
        L_{A} &= \sum_{(x,r^x)\in D_x,(y,r^y) \in D_y}-Cos(\theta_X(x),\theta_Y(y)) \: W(r^x,r^y) 
    \end{split}\\
    \begin{split}\label{eq:inter-w}
        W(r^x,r^y) &= exp\left(\frac{-|r^x-r^y|}{\delta^2}\right)
    \end{split}
\end{align}
where $Cos(\cdot,\cdot)$ is the cosine similarity, $W(\cdot,\cdot)$ a function that computes the similarity coefficient given a pair of rewards, $\delta$ a constant that controls how steep the similarity is with respect to the reward difference, and $r^x,r^y$ the immediate normalized rewards of $x,y$, respectively. Compared to Eq. \ref{eq:uma}, the alignment loss uses the cosine similarity to prevent mapping functions $\theta_X,\theta_Y$ from producing vectors with small magnitude (which would minimize a Euclidean-based loss). Thus, considering that the similarity coefficient grows as reward difference shrinks, Eq. \ref{eq:alignment} penalizes mapping states/actions apart if they yield similar rewards.

To maintain, in the latent space, the local neighborhood relations present within the aligned domains, Eq. \ref{eq:geo-pre} describes the geometry preserving loss:

\begin{align}
    \begin{split}\label{eq:geo-pre}
        L_{G} = &\sum_{(x_i,r_i^x),(x_j,r_j^x)\in D_x} \frac{[Cos_d(x_i,x_j)-Cos(\theta_X(x_i),\theta_X(x_j))]^2}{|D_x|^2 - |D_x|} \: + \\
        &\sum_{(y_i,r_i^y),(y_j,r_j^y)\in D_y} \frac{[Cos_d(y_i,y_j)-Cos(\theta_Y(y_i),\theta_Y(y_j))]^2}{|D_y|^2 - |D_y|}
    \end{split}\\
    \begin{split}\label{eq:cos-ned}
        Cos_d(a,b) = &-2 \cdot d(a,b) + 1
    \end{split}\\
    \begin{split}\label{eq:ned}
        d(a,b) = &\frac{\sqrt{\sum_i^N(a_i - b_i)^2}}{\sqrt{N}}
    \end{split}
\end{align}
where $N$ is the dimensionality of vectors $a,b \in \mathbb{R}^N$, $d(\cdot,\cdot)$ is the normalized Euclidean distance between two vectors, and $Cos_d(\cdot,\cdot):[0,1] \rightarrow [-1,1]$ is a linear mapping from the normalized Euclidean distance to the range of the Cosine similarity. By normalizing the Euclidean distance between the normalized vectors (Eq. \ref{eq:ned}), it is possible to linearly map the Euclidean distance to the Cosine similarity (Eq. \ref{eq:cos-ned}). Thus, with a reference available, the geometry preserving loss (Eq. \ref{eq:geo-pre}) penalizes mapping vectors too far (or too close) in the latent space, given their Euclidean distance in the input space.

To measure the similarity and transfer knowledge across tasks, it is necessary to map states and actions back and forth. Therefore, in addition to the alignment functions $\theta_X,\theta_Y$, a pair of functions $\phi_X : \mathbb{R}^m \rightarrow \mathbb{R}^p,\phi_Y : \mathbb{R}^m \rightarrow \mathbb{R}^q$ are trained to learn the inverse mapping of their respective counterpart, as well to remain consistent with respect to the other domain. Both objectives are modeled by Eq. \ref{eq:rec-xy} and Eq. \ref{eq:cyc-con-xy}, respectively:

\begin{align}
    \begin{split}\label{eq:rec-xy}
        L_{R_X} &= \left( x - \phi_X \circ \theta_X(x) \right)^2\\
        L_{R_Y} &= \left( y - \phi_Y \circ \theta_Y(y) \right)^2
    \end{split}\\
    \begin{split}\label{eq:cyc-con-xy}
        L_{C_X} &= \left( x - \phi_X \circ \theta_Y \circ \phi_Y \circ \theta_X(x) \right)^2\\ 
        L_{C_Y} &= \left( y - \phi_Y \circ \theta_X \circ \phi_X \circ \theta_Y(y) \right)^2
    \end{split}
\end{align}
where $\circ$ is the function composition operator. The reconstruction losses (Eq. \ref{eq:rec-xy}) promote retrieving the input vectors back from the latent space, while the cycle consistency losses (Eq. \ref{eq:cyc-con-xy}) \cite{gupta2017learning} encourage consistency across spaces. Thus, the Reward-Based Alignment (ReBA) loss is described by Eq. \ref{eq:rebal}:

\begin{align}
    \begin{split}\label{eq:rebal}
        L_{ReBA} &= \lambda_1 \cdot L_{A} + \lambda_2 \cdot L_{G} + \lambda_3 \cdot (L_{R_X} + L_{R_Y}) \\
        &+ \lambda_4 \cdot (L_{C_X} + L_{C_Y}) + \lambda_5 \cdot (\norm{\theta_X(x)}_2 + \norm{\theta_Y(y)}_2)
    \end{split}
\end{align}
where $\lambda_1,\lambda_2,\lambda_3,\lambda_4,\lambda_5$ are constant importance weights (which are determined empirically), and $\norm{\theta_X(x)}_2 + \norm{\theta_Y(y)}_2$ the L2-norm of the latent vectors, used as regularizing terms to promote vectors with smaller magnitudes, and improve the stability of the training process.

\subsection{Alignment Learning}\label{sec:learn-rebal}
\begin{figure}
    \centering
    \includegraphics[width=\textwidth]{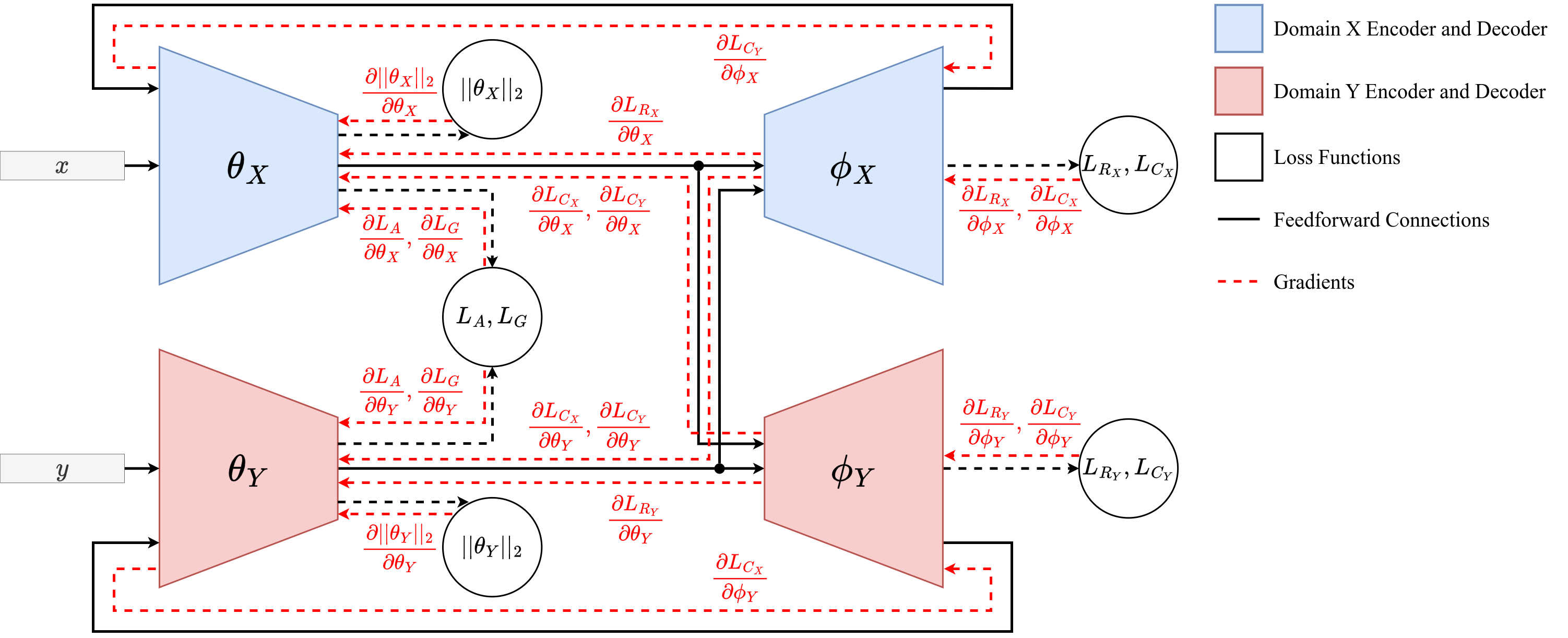}
    \caption{A pair of encoders ($\theta_X,\theta_Y$) are trained to minimize the alignment (Eq. \ref{eq:alignment}), geometry preserving (Eq. \ref{eq:geo-pre}) and regularization losses, while the encoder coupled with the decoders ($\phi_X,\phi_Y$) optimize the reconstruction (Eq. \ref{eq:rec-xy}) and cycle-consistency losses (Eq. \ref{eq:cyc-con-xy}). The solid black and dashed red lines describe the feed-forward and gradient back propagation, respectively.}
    \label{fig:training-losses}
\end{figure}

The alignment between each pair of spaces is learned independently of each other (see Fig. \ref{fig:training-losses}). Within each training process, a pair of encoders map vectors from their original spaces to a common latent space, and are coupled by the alignment (Eq. \ref{eq:alignment}) and geometry preserving losses (Eq. \ref{eq:geo-pre}). On the other hand, the decoding models strive to map latent vectors to equivalent observations from the original and other domain, through the minimization of the reconstruction (Eq. \ref{eq:rec-xy}) and cycle-consistency losses (Eq. \ref{eq:cyc-con-xy}), respectively.

Let $D_X=\{( s_i^x,a_i^x,r_i^x ) | ( s_i^x,a_i^x,r_i^x,s^x) \thicksim MDP_X \}$, $D_Y=\{( s_j^y,a_j^y,r_j^y ) | ( s_j^y,a_j^y,r_j^y,s^y) \thicksim MDP_Y \}$ be a pair of normalized state-action-reward data sets sampled from MDPs $MDP_X$ and $MDP_Y$. Then, the state and action alignment data sets ($D_{Sx},D_{Sy}$ and $D_{Ax},D_{Ay}$) are defined as follows:

\begin{align*}
    &D_{Sx} = \{ (x_i,r_i^x) \: | \: x_i=s_i^x, ( s_i^x,a_i^x,r_i^x ) \in D_X \} \\
    &D_{Sy} = \{ (y_j,r_j^y) \: | \: y_j=s_j^y, ( s_j^y,a_j^y,r_j^y ) \in D_Y \} \\
    &D_{Ax} = \{ (x_i,r_i^x) \: | \: x_i=a_i^x, ( s_i^x,a_i^x,r_i^x ) \in D_X \} \\
    &D_{Ay} = \{ (y_j,r_j^y) \: | \: y_j=a_j^y, ( s_j^y,a_j^y,r_j^y ) \in D_Y \}
\end{align*}
That is, although the two alignment training processes are performed independently, the reward is shared by the state and action that were observed along with it.

\paragraph{Similarity-based Data Pair Matching}
Given the definition of the alignment loss (Eq. \ref{eq:alignment}), it is necessary to map every data point from both domains to the latent space to compute the loss for a single pair of data points ($x,y$). Sometimes in practice, such computational requirements exceed available resources, therefore, an approximation is necessary.

Considering that, for pairs of latent vectors with non-zero distances between them, the alignment loss is a monotonic increasing function of the similarity coefficient, the data pairs with more similar rewards are those that cause a greater change in the model's weights. To mitigate the computational burden of computing the alignment loss, an approximation of Eq. \ref{eq:alignment} is presented in Eq. \ref{eq:alignment-approx}:

\begin{align}
    \begin{split}\label{eq:alignment-approx}
        L_{A} \approxeq& \thicktilde{L_{A}} = \sum_{(x,r^x,y,r^y) \in D(D_x,D_y)}-Cos(\theta_X(x),\theta_Y(y)) \: W(r^x,r^y)
    \end{split}
\end{align}
\begin{align*}
    D(D_x,D_y) =& \{ (x,r^x,y,r^y) \: | \: \forall (x,r^x) \in D_x, (y,r^y)=\argmax_{(y_i,r_i^y)} W(r^x,r_i^y) \} \: \cup \\
    &    \{ (x,r^x,y,r^y) \: | \: \forall (y,r^y) \in D_y, (x,r^x)=\argmax_{(x_i,r_i^x)} W(r_i^x,r^y) \}
\end{align*}
where $W(r^x,r^y)$ is the similarity coefficient  (Eq. \ref{eq:inter-w}) and $D(\cdot,\cdot)$ a function that returns a subset of the training data set, in which every data point is paired with the most similar match from the other domain. Thus, to train the state and action alignments with the approximated loss (Eq. \ref{eq:alignment-approx}), the training data sets would be $D(D_{Sx},D_{Sy})$ and $D(D_{Ax},D_{Ay})$, respectively.

\paragraph{Geometry Preserving Neighborhood Selection}
Similarly to how data pairs are matched based on their reward-similarity, the set of data points used to compute the geometry preserving loss (Eq. \ref{eq:geo-pre}) can be filtered to reduce the computational workload. Instead of computing the distance between every pair in a data set, the approximated loss with a subset of neighbors, as described by Eq. \ref{eq:geo-pre-approx}:

\begin{align}
    \begin{split}\label{eq:geo-pre-approx}
        L_{G} \approxeq \thicktilde{L_{G}} = &\sum_{(x_i,r_i^x)\in D_x} \sum_{(x_j,r_j^x) \in K(x_i,D_x,k)} \frac{[Cos_d(x_i,x_j)-Cos(\theta_X(x_i),\theta_X(x_j))]^2}{|D_x| \cdot k} \: + \\
        &\sum_{(y_i,r_i^y) \in D_y} \sum_{(y_j,r_j^y) \in K(y_i,D_y,k)} \frac{[Cos_d(y_i,y_j)-Cos(\theta_Y(y_i),\theta_Y(y_j))]^2}{|D_y| \cdot k}
    \end{split}
\end{align}
where $Cos_d(\cdot,\cdot)$ is defined by Eq. \ref{eq:cos-ned}, $k$ the number of neighbors to include in each data point's geometry preserving computation, and $K(x,D,k)$ a function that returns the $k$ nearest neighbors (using the Euclidean distance) of $x$ in $D$.

\section{Similarity-based Knowledge Transfer}\label{sec:cross-domain}
\begin{figure}
    \centering
    \includegraphics[width=\textwidth]{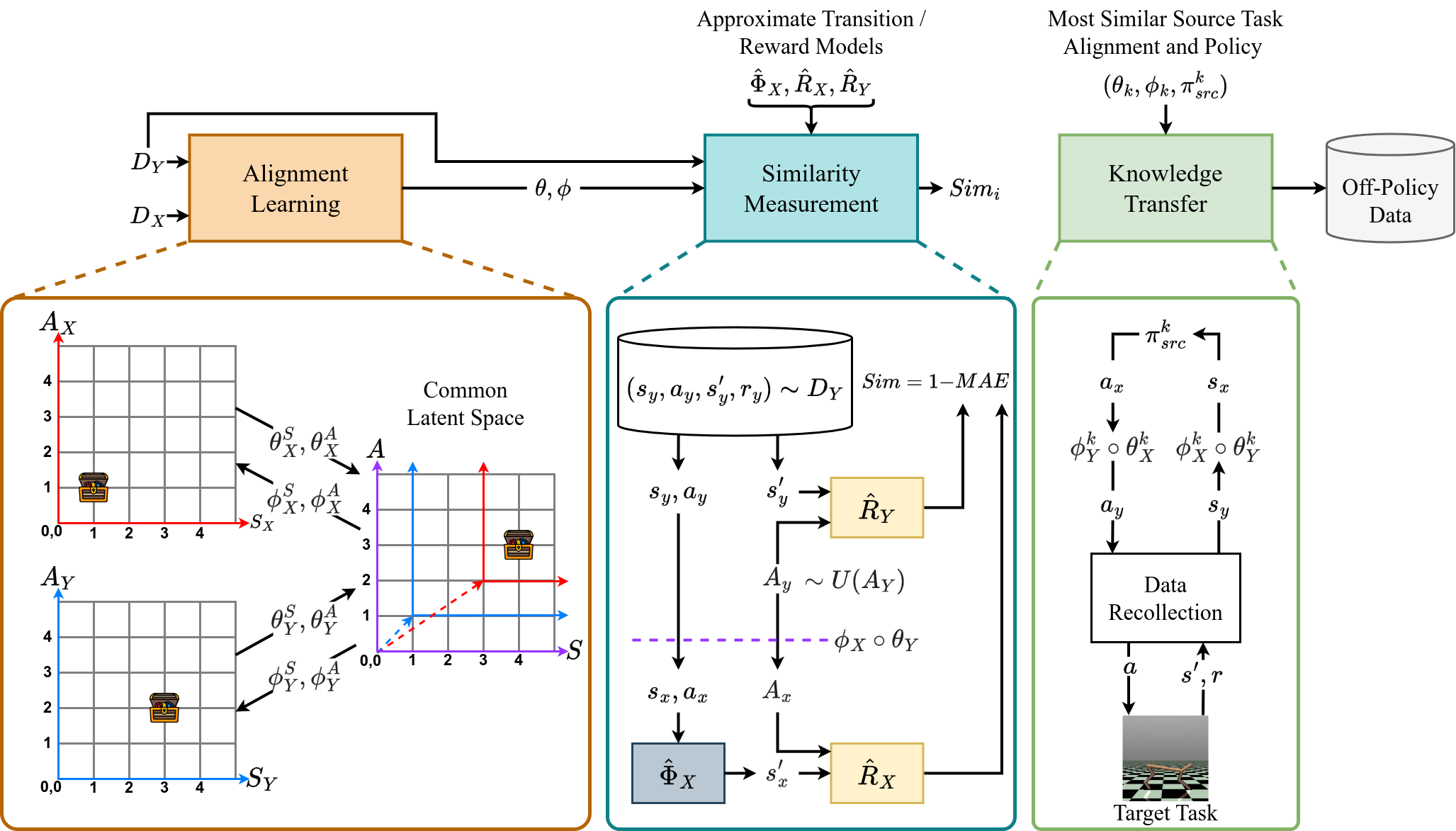}
    \caption{Interaction between the three main processes of \textit{SimKnoT}: alignment learning, similarity measurement, and knowledge transfer. After learning the alignment models (with source and target data sets $D_X,D_Y$), target state-actions are mapped to the source domain so that the reward distribution of the target and source next state ($s'_y,s'_x$) are compared to compute the source-target similarity. Finally, the policy and alignment models from the most similar source task are used to sample an off-policy data set that the RL algorithm can use to learn faster.}
    \label{fig:method-detail}
\end{figure}

Once the alignment functions (and their inverse) have been learned, the set of learned state and action encoders-decoders can be used to compare the reward and transition dynamics of a pair of RL tasks (Section \ref{sec:task-simi}). Then, the most similar task will be used as the source of knowledge to transfer actions to the target learner and accelerate the exploration process (Section \ref{sec:knowledge-transfer}), see Fig. \ref{fig:method-detail}.

\subsection{Inter-Task Similarity}\label{sec:task-simi}
Let $D_X,D_Y$ be data sets sampled from source and target tasks $T_X=\langle S_X,A_X,D_X,\pi_X \rangle$, $T_Y = \langle S_Y,A_Y,E_Y \rangle$, and $(\theta_X^S,\theta_X^A,\theta_Y^S,\theta_Y^A)$, $(\phi_X^S,\phi_X^A,\phi_Y^S,\phi_Y^A)$ the encoders and decoders that align their state and action spaces, respectively. Given the approximate models:
\begin{align*}
    \begin{split}
        &\hat{R_X} : S_X \times A_X \rightarrow \mathbb{R} \\
        &\hat{R_Y} : S_Y \times A_Y \rightarrow \mathbb{R}
    \end{split}
\end{align*}
trained with data sets $D_X,D_Y$ to predict the immediate reward in the source and target tasks, and $\hat{\Phi_X} : S_X \times A_X \rightarrow S_X$ trained with $D_X$ to predict the next state in the source task. Then, the inter-task similarity between tasks $T_X$ and $T_Y$ is given by Eq. \ref{eq:task-sim}:
\begin{align}
    \begin{split}\label{eq:task-sim}
    Sim(T_X,T_Y) &= 1 - \sum_{s_y,a_y,s'_y,r_y\in D_Y} \sum_{a'_x \in A_x, a'_y \in A_y} \frac{|\hat{R_X}(s'_x,a'_x) - \hat{R_Y}(s'_y,a'_y)|}{ |D_Y| \cdot |A_x| }  \\
    s'_x &= \hat{\Phi_X}(\phi_X^S \circ \theta_Y^S(s_y),\phi_X^A \circ \theta_Y^A(a_y)) \\
    A_y &\thicksim U(A_Y) \\
    A_x &= \phi_X^A \circ \theta_Y^A(A_y)
    \end{split}
\end{align}
where $s'_x$ is the next-state prediction according to domain $X$'s transition model, given a state and action mapped from domain $Y$, $A_y$ a set of domain $Y$ actions sampled from a uniform distribution, and $A_x$ their mapping to domain $X$'s action space.

Equation \ref{eq:task-sim} computes high similarity scores for tasks when their transition models change the environment's to states with similar reward distributions, given that the action and previous state are equivalent, according to the alignment functions. In other words, the similarity function compares one-step transitions that lead to states from which a random agent can expect similar rewards.

\subsection{Knowledge Transfer}\label{sec:knowledge-transfer}
To transfer knowledge from a set of source tasks, the \textbf{\textit{Sim}}ilarity-based \textbf{\textit{Kno}}wledge \textbf{\textit{T}}ransfer (\textit{SimKnoT}) algorithm uses the similarity function to select the best candidate, among a pool of source candidates, to transfer optimal-policy actions (see Algorithm \ref{alg:transfer-rl}). Given a set of source tasks ($T_{src}$) and a target task ($T_{tgt}$), the following five-step process is performed:
\begin{enumerate}
    \item Perform standard RL and store observations in replay buffer $B$ and data set $D$.
    \item Use data set $D$ to learn the alignment models between $T_{tgt}$ and every source task.
    \item Transfer actions from the most similar source task' policy (via the alignment models).
    \item Optimize the RL objective over the transitions obtained with the source policy.
    \item Perform standard RL to finish the training of $\pi_{tgt}$.
\end{enumerate}

That is, the goal of \textit{SimKnoT} is to provide a kickstart boost, with off-policy data, and reduce the sample complexity of the baseline RL algorithm. Additionally, the extra optimization steps, taken after knowledge is transferred (lines 13-15 in Algorithm \ref{alg:transfer-rl}), help the target policy to assimilate the effect of the off-policy data before it dilutes with future on-policy observations.

\begin{algorithm}
\caption{$SimKnoT$: Similarity-based Knowledge Transfer}\label{alg:transfer-rl}
\begin{algorithmic}[1]
\Require $T_{src}=\{ T_{src}^1 ,..., T_{src}^N \}, T_{tgt}=\langle S_{tgt}, A_{tgt}, E_{tgt} \rangle$
\Ensure $\pi_{tgt}$
\LeftComment{Train in RL setting and sample data}
\State $D \gets \{\}, \: B \gets \{\}, \: \pi_{tgt} \gets InitPolicy()$
\For{$i \gets 1$ to $I_{preTL}$}\\
    $\tab\text{Sample transition }(s,a,s',r)\text{ with }(\pi_{tgt},E_{tgt})\text{ and add it to }B\text{ and }D$\\
    $\tab\text{Optimize RL objective of }\pi_{tgt}\text{ with respect to }B$
\EndFor
\Statex
\LeftComment{Learn alignment and measure similarity}
\State $Sim \gets [\:], \: Models \gets [\:]$
\For{$i \gets 1$ to $N$}\\
    $\tab\text{Optimize Eq. \ref{eq:rebal} with }(D,GetData(T_{src}^i))\text{ and add the alignment}$
    $\tab\text{models to }Models$\\
    $\tab\text{Compute similarity (Eq. \ref{eq:task-sim}) with }(D,GetData(T_{src}^i))\text{ and add the}$
    $\tab\text{similarity score to }Sim$
\EndFor
\State Get index of most similar task from $Sim$ and store it in $simID$
\Statex
\LeftComment{Sample data with transferred policy and update target policy}
\State $\pi_{tgt},\: B \gets TransferPolicy(T_{src},Models,simID,\pi_{tgt},B,E_{tgt})$
\Statex
\LeftComment{Update $\pi_{tgt}$ without adding data to the replay buffer}
\For{$i \gets 1$ to $E$}\\
    $\tab\text{Optimize RL objective of }\pi_{tgt}\text{ with respect to }B$
\EndFor
\Statex
\LeftComment{Train in RL setting}
\For{$i \gets 1$ to $I_{posTL}$}\\
    $\tab\text{Sample transitions }(s,a,s',r)\text{ with }(\pi_{tgt},E_{tgt})\text{ and add it to }B$
    $\tab\text{Optimize RL objective of }\pi_{tgt}\text{ with respect to }B$
\EndFor
\\
\Return $\pi_{tgt}$
\end{algorithmic}
\end{algorithm}

\begin{algorithm}
\caption{Source Policy Transfer to Target Domain}\label{alg:transfer-pol}
\begin{algorithmic}[1]
\Require $T_{src},Models,simID,\pi_{tgt},B,E_{tgt}$
\Ensure $\pi_{tgt}, B$
    \Function{TransferPolicy}{$T_{src},Models,simID,\pi_{tgt},B,E_{tgt}$}
        \Statex $\tab\triangleright\text{ Source policy and alignment models}$
        \State $\pi_{src} \gets GetPolicy(T_{src}^{simID})$
        \State $\phi^S_{src},\theta^S_{tgt},\phi^A_{tgt},\theta^A_{src} \gets GetAlignmentModels(Models[simID])$
        \Statex
        \Statex $\tab\triangleright\text{ Perform off-policy RL with source policy}$
        \For{$i \gets 1$ to $I_{TL}$}
            \State $\text{Sample }(s,a_{tgt},s',r)\text{ from }E_{tgt}\text{, where }a_{src} = \pi_{src}(\phi^S_{src} \circ \theta^S_{tgt}(s_{tgt}))$
            $\tab\tab a_{tgt} = \phi^A_{tgt} \circ \theta^A_{src}(a_{src})\text{, and add it to }B$
            \State$\text{Optimize RL objective of }\pi_{tgt}\text{ with respect to }B$
        \EndFor
        \\
        \Return $\pi_{tgt}, B$
    \EndFunction
\end{algorithmic}
\end{algorithm}

\section{Experiments}\label{sec:exp}
In order to evaluate the ability of \textit{SimKnoT} to work as a knowledge transfer method for the latent unobservable scenario (see Sec. \ref{sec:its}),
we present a set of experiments to answer the following research questions:
\begin{enumerate}
    \item Can the similarity measure identify inter-task similarity in the cross-domain setting?
    
    \item Does \textit{SimKnoT} benefit the target learner?

    \item Can \textit{SimKnot} perform positive transfer among tasks with no semantic relation?

    \item Does the transferred knowledge impact the target learner's performance?
\end{enumerate}

The set of evaluation tasks is constituted by six Mujoco control tasks \cite{todorov2012mujoco} from the Gymnasium suite \cite{towers_gymnasium_2023}: half cheetah (H. Cheetah), hopper, inverted pendulum (IP), swimmer, walker2D and inverted double pendulum (IDP), from which the IP-IDP and hopper-walker2D pairs are considered to belong to the latent-observable setting (see Sec. \ref{sec:its}) given the shared high-level structure. Additionally, considering our work's dense-reward assumption (see Sec. \ref{sec:scope}), the original IP reward function has been changed from returning $1.0$ each step the pole is standing to the following form:
\begin{equation}
    R_{IDP\:Dense}(\theta,\dot{\theta}) = C_{Alive} - 0.01 \cdot sin(\theta)^2 - (cos(\theta)-1)^2 - 0.005 \cdot \dot{\theta}^2
\end{equation}
where $C_{Alive}$ is a bonus for staying alive, the $sin,cos$ terms penalize the angular distance between the pole and the perfect vertical position, whereas the last term penalizes the angular velocity of the pole. The constant values have been copied from the original IDP task.

For comparison purposes, in addition to \textit{SimKnoT}, we used the MushroomRL \cite{mushroomrl} implementation of the Soft Actor-Critic (SAC) \cite{haarnoja2018soft} algorithm, as well as a modification which we refer to as SAC with fixed-buffer optimization (SAC+FBO), in which at some point of the training process the algorithm updates the actor-critic (for a fixed number of steps) without adding data to the replay buffer (see lines 13-15 in Algorithm \ref{alg:transfer-rl}), as \textit{SimKnoT} does. The cumulative reward (averaged over 20 episodes) of a policy trained with 10 million observations (10M expert) and a random policy are shown to contrast the performance of the learning agents. Regarding the multi-source transfer work proposed by \cite{heng2022crossdomain} (closest work to ours), we did not compare against their method because to compute the \textit{correction loss} (Eq. 4 in \cite{heng2022crossdomain}) the target environment must be differentiable, so that the loss gradient can reach the action encoders, which is a requirement our experimental setting can not meet.

Experiments consist of learning the alignment models and measuring similarity between a target task and each source task, followed by the knowledge being transferred from the most similar task. Beforehand, a SAC agent has been trained in each source task, for 1 million (hopper, IP, IDP) and 3 million (H. cheetah, swimmer, walker2D) environment steps, resulting in a trained policy and a data set of $(state, action, next state, reward)$ observations, which is used to train the transition and reward approximation models ($\hat{\Phi_X},\hat{R_X}$ in Sec. \ref{sec:task-simi}).
The target task data set consists of the first 50,000 observations sampled by the SAC algorithm, which are used to learn the target reward approximation model ($\hat{R_Y}$ in Sec. \ref{sec:task-simi}), whereas the alignment models learning and similarity measurements are computed with the source-target data set pair.

When the most similar task is identified, through the alignment models, actions are transferred from the source policy for the learning agent to perform in the target environment for 100,000 environment steps. Then, both \textit{SimKnoT} and SAC+FBO perform 4,000 learning updates on their respective agent without adding observations to the replay buffer. For more details of our experimental setting, see Appendix \ref{appendix}.

\subsection{Experiment Results}
\begin{figure}
    \centering
    \includegraphics[width=0.8\linewidth]{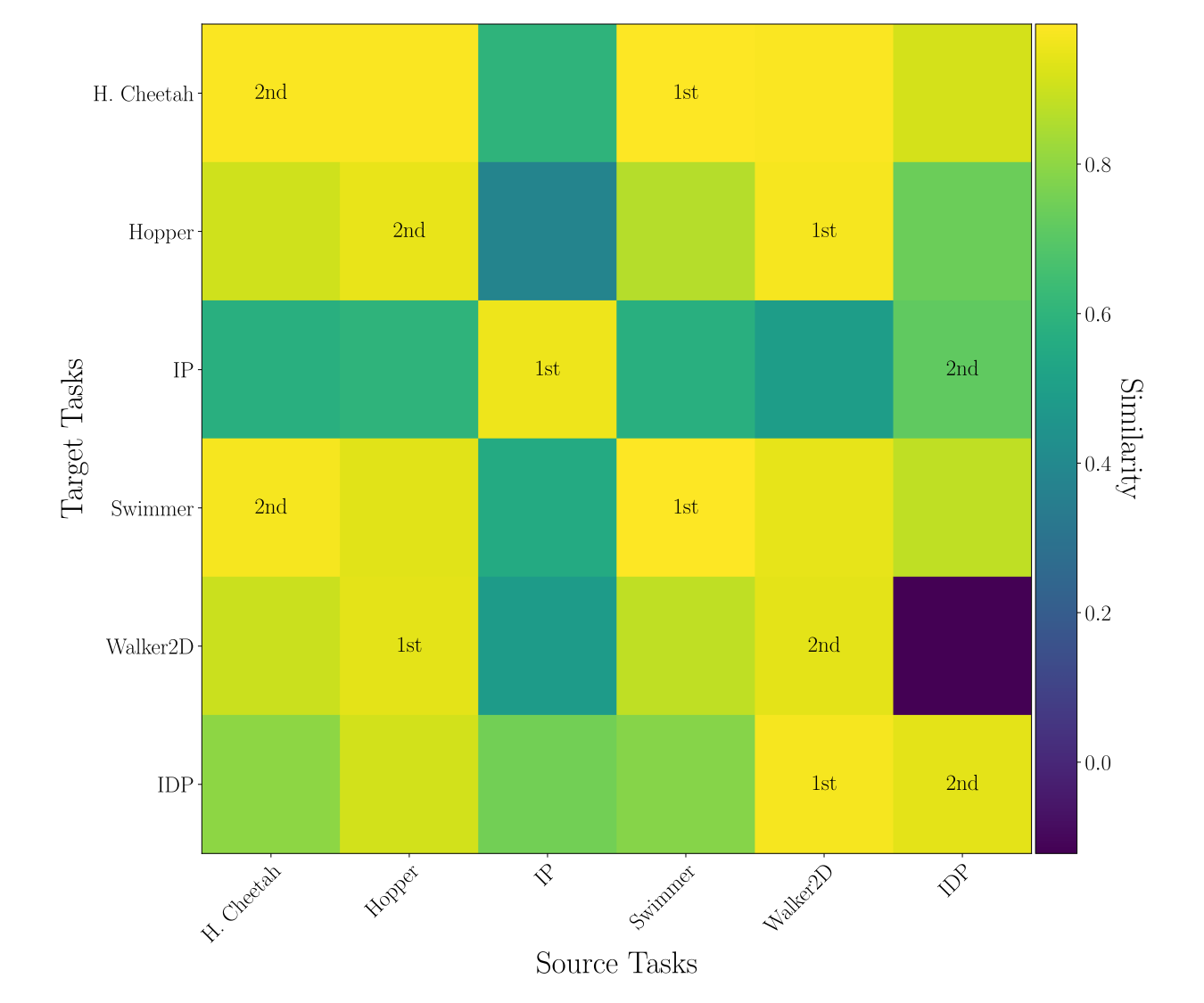}
    \caption{Similarity matrix computed with Eq. \ref{eq:task-sim}, where the higher the value, the more similar two tasks are. Rows and columns represent target and source tasks, respectively. The first and second most similar source task to each target task are specified with a label.}
    \label{fig:sim-mat-all}
\end{figure}

Figure \ref{fig:sim-mat-all} shows the similarity scores obtained after computing Eq. \ref{eq:task-sim} between each pair of source (column) and target (row) task, using 50,000 observations from the target task, where two tasks (\textit{i.e.} IP, Swimmer) out of six were successfully identified as the most similar with their source counterpart. On the other hand, the remaining four tasks (H. Cheetah, Hopper, Walker2D, IDP) ranked their source twin as the second most similar task.

\begin{figure}
    \centering
    \includegraphics[width=\linewidth]{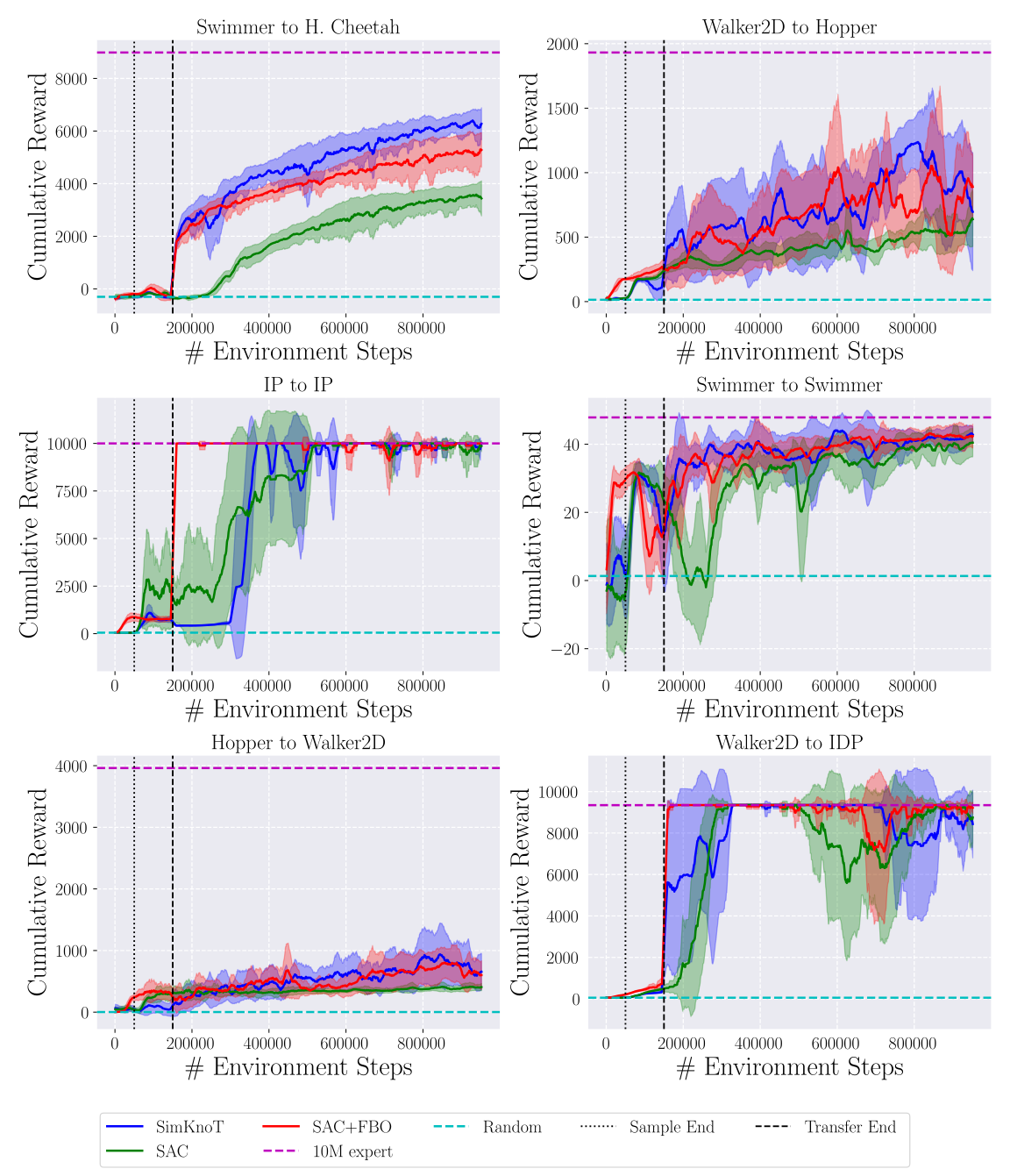}
    \caption{Episodic cumulative reward (averaged over 5 trials) in target tasks of \textit{SimKnoT} (transferring from the most similar task according to Fig. \ref{fig:sim-mat-all}), SAC and SAC+FBO. The vertical dotted line shows the timestamp at which \textit{SimKnoT} stopped sampling data for aligning/measuring purposes, the space between the dotted and dashed shows its transfer period, while the dashed displays the moment at which \textit{SimKnot} and SAC+FBO performed the fixed-buffer optimization. Top and bottom horizontal lines show the average performance of a policy trained with 10 million observations and a random policy.}
    \label{fig:transfer-all}
\end{figure}

Regarding the knowledge transfer results, Fig. \ref{fig:transfer-all} shows the performance of \textit{SimKnoT}, SAC and SAC+FBO in each target task, averaged over 5 trials and smoothed with a uniform and symmetrical centered window of length 15. In the case of \textit{SimKnoT}, the reported results correspond to transferring knowledge from the most similar source task, according to the similarity scores in Fig. \ref{fig:sim-mat-all}. Additionally, Table \ref{tab:tl-performance} summarizes the top performance, and the number of training environment steps at which it was reached, of all three learning methods. To avoid outlier values, the reported performances are the highest cumulative reward sustained for at least 10 consecutive evaluations (which is equivalent to 10,000 environment steps.)

\paragraph{Can the similarity measure identify inter-task similarity in the cross-domain setting?}
Considering that the purpose of the similarity measure is to work as a heuristic to select the best source of knowledge available for transfer purposes, and in view of the knowledge transfer results (see Fig. \ref{fig:transfer-all}) in which in five out of six times \textit{SimKnot} obtains a better performance than the baseline SAC agent after transferring knowledge from the most similar task, it is safe to say that the similarity measure can detect inter-task similarity. 

On the other hand, from a qualitative perspective, the similarity measure shows certain results worth mentioning. For instance, despite ranking a task with itself as the most similar in only two cases (IP and Swimmer), it is also able to identify a top-two similarity between IP-IDP and Hopper-Walker2D. This result is particularly interesting because it is akin to the task pairing some works use as an example of tasks with different morphology, but that are likely to benefit from transferring knowledge due to their high-level similarity.

\paragraph{Does SimKnoT benefit the target learner?}
As shown in Fig. \ref{fig:sim-mat-all}, the similarity measure is able to rank every source task as at least the second most similar to its target counterpart, despite the variety of available options. However, despite the source-target matching being far from perfect, we are able to transfer knowledge in a way that helps the target learner to reach larger rewards with fewer observations, in comparison to SAC. The knowledge transfer process resulted in \textit{SimKnoT} outperforming SAC in 5 (H. Cheetah, Hopper, IP, Swimmer, Walker2D) out of 6 tasks (see Table \ref{tab:tl-performance}).

On the other hand, the Hopper-Walker2D, Walker2D-Hopper and IDP-IP (in second place) source-target matching made by the similarity measure are good examples of latent observable pairs. Similar to how some works \cite{heng2022crossdomain,zhang2021learning} transfer knowledge across tasks with known shared latent structures (\textit{e.g.} similar morphology with different number of links/limbs), our method is capable of overcoming the representation mismatch to reuse knowledge. However, there exists an important difference to \textit{SimKnoT}, as the similarity measure endows it with the ability to detect such inter-task relations without the supervision of an expert.

\paragraph{Can SimKnot perform positive transfer among tasks with no semantic relation?}
In addition to reusing knowledge across tasks with shared high-level features (\textit{e.g.} Hopper and Walker2D), by transferring knowledge from Swimmer (a 3-limb crawling robot) to H. Cheetah (a 9-limb dog like robot) \textit{SimKnoT} shows its ability to improve the performance of a target learner, despite the lack of discernible commonalities as H. Cheetah not only deals with inertial forces but also with the constant effect of gravity (see the upper-left graph in Fig. \ref{fig:transfer-all}).

Albeit we are not the first work to achieve positive transfer in the latent unobservable setting (\textit{e.g.} \cite{raychaudhuri2021cross} successfully transfer knowledge from the Mujoco Ant task to H. Cheetah), to the best of our knowledge, \textit{SimKnoT} is the first method to select a non-related source from a varied set of options, and produce positive transfer in the target domain. By doing so, the proposed method offers an option to automate the selection of source tasks in settings where an expert can not provide supervision due to the sheer number of problems, \textit{e.g.} reducing the data required to solve every instance in a database of RL tasks \cite{ramos2021rlds} by reusing data when possible.

\paragraph{Does the transferred knowledge impact the target learner's performance?}
Considering that in Deep RL policies and value functions are modeled by neural networks, it is possible to improve the model's performance by solely increasing the number of weight updates with the same data set, as \textit{SimKnoT} does after transferring actions from the source policy (see lines 13-14 in Algorithm \ref{alg:transfer-rl}). Thus, to evaluate the impact of the data recollected by \textit{SimKnoT} in the target learner, we compare its performance against SAC+BFO, which is a SAC agent that performs the same number of weight updates (4,000) as \textit{SimKnoT}, at the same time (after 150,000 environment steps).

As shown in Fig. \ref{fig:transfer-all} and Table \ref{tab:tl-performance}, \textit{SimKnoT} achieves a larger cumulative reward in fewer environment steps than SAC+FBO in 4 (H. Cheetah, Hopper, Swimmer, Walker2D) out of 6 tasks. This shows that it is not only the increased number of learning updates, but also the data set what causes the performance improvement of \textit{SimKnoT}. Regarding the two tasks in which SAC+FBO outperforms the other methods, both IP and IDP are the only tasks whose cap performance can be achieved in less than 1 million environment steps. This is an example of how, in less complex tasks, it is more difficult for knowledge transfer methods to offer a benefit, as simpler solutions are likely to display competitive performances.

\begin{table}[]
\caption{Top performance comparison of SAC, SAC+FBO and \textit{SimKnoT} in each target task (averaged over 5 trials), as well as the sample complexity required to achieve such performance. The reported top cumulative rewards consist of the highest performance sustained by the methods for at least 10,000 consecutive environment steps. The larger the reward and the fewer environment steps, the better (the best performance of each task is in bold font).}
\label{tab:tl-performance}
\centering
\begin{tabular}{ccccccc}
\hline
\multirow{2}{*}{Method}  & \multicolumn{6}{c}{Top Cumulative Reward $\uparrow^+$ / Env. Steps $\downarrow^+$ } \\ \cline{2-7} 
                         & H. Cheetah   & Hopper   & IP        & Swimmer  & Walker2D  & IDP  \\ \hline
\multirow{2}{*}{SAC}     & 3,535 /      & 570 /    & 9,996 /   & 39.86 /  & 405.35 /  & 9,353 /  \\
                         & 939,000      & 951,000  & 547,000   & 951,000  & 932,000   & 330,000    \\
\multirow{2}{*}{SAC+FBO} & 5,204 /      & 980 /    & \textbf{9,996 /}   & 42.83 /  & 782.84 /  & \textbf{9,348} /  \\
                         & 918,000      & 849,000  & \textbf{173,000}   & 850,000  & 862,000   & \textbf{187,000}    \\
\multirow{2}{*}{SimKnoT} & \textbf{6285 /}       & \textbf{1,222 /}  & 9,996 /   & \textbf{43.66 /}  & \textbf{885.45 /}  & 9,352 /  \\
                         & \textbf{932,000}      & \textbf{812,000}  & 383,000   & \textbf{605,000}  & \textbf{853,000}   & 341,000    \\ \hline
\end{tabular}
\end{table}

\subsection{Ablation Studies}
In order to determine the importance of each loss term proposed to learn the cross-domain alignment, Table \ref{tab:ablation} summarizes the similarity score when IDP is both the source and target task. In all the experiments, the source and target data sets had a size of 1 million and 50,000 observations, respectively. Results show how the similarity score slightly decreases when either the alignment or geometry-preserving loss is removed, and how it plummets when both terms are not used to train the alignment models. Thus, it is safe to argue that both terms play a crucial role in the unsupervised selection of source tasks, which in turn is the cornerstone of \textit{SimKnoT}.

\begin{table}
\caption{Comparison of the alignment training setting (with IDP as source and target task) used in the transfer experiments (first row) and multiple variations of ReBA without the alignment (w/o A) and geometry preserving (w/o GP) terms. The rightmost column shows the similarity score as defined in Eq. \ref{eq:task-sim}.}
\label{tab:ablation}
\centering
\begin{tabular}{cc}
\hline
Loss    & Similarity       \\ \hline
ReBA               & 0.9401           \\
ReBA w/o A               & 0.9024           \\
ReBA w/o GP              & 0.9305           \\
ReBA w/o A, GP  & -0.2037 \\ \hline
\end{tabular}
\end{table}

\section{Conclusions}\label{sec:conclusions}
We have presented a method to measure inter-task similarity, and to transfer knowledge, across tasks that do not share representation or have common high-level features. This research introduced a semi-supervised method to align state-action spaces based on the immediate reward without any special data recollection assumptions, a similarity measure able to identify similar tasks (for transfer purposes) in highly diverse sets, and a knowledge transfer method capable of 
 outperforming baseline RL methods. Moreover, in addition to being the first method to automate the selection of sources of knowledge without making assumptions of inter-task similarity in the cross-domain setting, \textit{SimKnoT} can easily be embedded in any RL algorithm that supports using off-policy data.

Among the main limitations of our work, we find the dense reward assumption to be the most restrictive one. As tasks with sparse rewards are still a challenging problem, it is important to devise methods that work in settings with scarce feedback, which are commonly found in real-world applications. Moreover, assuming normalized rewards is another prohibitive limitation to our work, as not knowing how the current experience of the learning agent compares to future observations is a core element of the exploratory process in RL.

Finally, as future work, we would like to continue increasing the autonomy of the proposed method, by developing criteria to decide when the agent should end the sampling and transferring stages, as well as to identify if none of the source tasks will yield a positive transfer, so that the target learner avoids the negative transfer and learns from scratch instead. With these improvements, \textit{SimKnoT} would be able to take on even more challenging scenarios.

\paragraph{}
\textbf{Acknowledgements.} Sergio A. Serrano, with CVU 853275, thanks CONAHCYT for the PhD scholarship without which this research would not have been possible, as well as M.S. David Carrillo and the INAOE Robotics Laboratory which were important for the realization of this work.

\section*{Declarations}
\textbf{Funding}: This work was supported with PhD scholarship provided by CONAHCYT.\\
\textbf{Conflicts of interest/Competing interests}: The authors have no conflicts of interest, nor competing interests, to declare that are relevant to the content of this article.\\
\textbf{Ethics approval}: Not applicable.\\
\textbf{Consent to participate}: Not applicable.\\
\textbf{Consent for publication}: Not applicable.\\
\textbf{Availability of data and material}: Not applicable.\\
\textbf{Authors' contributions}: Sergio A. Serrano, Jose Martinez-Carranza and L. Enrique Sucar contributed to the study conception, design, and implementation. All authors read and approved the final manuscript.\\
\textbf{Code availability}: \url{https://github.com/saSerrano/simknot}\\

\bibliography{bibliography}

\appendix

\newpage
\section{Appendix A}\label{appendix}
In Tables \ref{tab:models-rt}, \ref{tab:models-ed}, and \ref{tab:models-sac} one will find the parameters used to train the reward/transition approximation models, the alignment encoder-decoders and the RL agents, respectively. For more detail on our implementation, please refer to \url{https://github.com/saSerrano/simknot}.

\begin{table}[h]
\caption{Parameters used to train the neural networks that approximate the reward and transition models, and that are used to measure similarity, as described in Eq. \ref{eq:task-sim}. $|S|$ and $|Y|$ are the number of state and action variables, respectively.}
\label{tab:models-rt}
\centering
\begin{tabular}{ccc}
\hline
\multirow{2}{*}{Parameter} & \multicolumn{2}{c}{Value}              \\ \cline{2-3} 
                           & Reward Model     & Transition Model    \\ \hline
\# Output Units            & 1                & $|S|$                 \\
Output Act. Fun.           & \multicolumn{2}{c}{Linear}             \\
Hidden Layers              & \multicolumn{2}{c}{{[}64,64,64,64{]}}  \\
Hidden Act. Fun.           & \multicolumn{2}{c}{ReLU}               \\
\# Input Units             & \multicolumn{2}{c}{$|S| + |A|$}          \\
Loss Function              & \multicolumn{2}{c}{Mean Squared Error} \\
Optimizer                  & \multicolumn{2}{c}{Adam}               \\
Batch Size                 & \multicolumn{2}{c}{512}                \\
Epochs                     & \multicolumn{2}{c}{300}                \\
Learning Rate              & \multicolumn{2}{c}{0.001}              \\ \hline
\end{tabular}
\end{table}

\begin{table}[h]
\caption{Parameters used to train the encoder-decoder pairs that are used to align the state and action spaces. $|X|$ and $|Y|$ are the dimensionality of the two spaces that will be aligned.}
\label{tab:models-ed}
\centering
\begin{tabular}{ccc}
\hline
\multirow{2}{*}{Parameter}                                                 & \multicolumn{2}{c}{Value}                                                              \\ \cline{2-3} 
                                                                           & Encoder                                    & Decoder                                   \\ \hline
\# Input Units                                                             & $|X|$                                      & $max(|X|,|Y|)+1$                          \\
\# Output Units                                                            & $max(|X|,|Y|)+1$                           & $|X|$                                     \\
Output Act. Fun.                                                           & \multicolumn{2}{c}{Linear}                                                             \\
Hidden Layers                                                              & \multicolumn{2}{c}{{[}64,64,64,64{]}}                                                  \\
Hidden Act. Fun.                                                           & \multicolumn{2}{c}{ReLU}                                                               \\
Loss Function                                                              & \multicolumn{2}{c}{ReBA (Eq. \ref{eq:rebal})}                         \\
Loss Weights                                                               & \multicolumn{2}{c}{$\lambda_1=1,\lambda_2=1,\lambda_3=1,\lambda_4=0.5,\lambda_5=0.05$} \\
$\delta$ in $L_A$ (Eq. \ref{eq:alignment})                & \multicolumn{2}{c}{0.25}                                                               \\
$k$ in $\thicktilde{L_{G}}$ (Eq. \ref{eq:geo-pre-approx}) & \multicolumn{2}{c}{2}                                                                  \\
Optimizer                                                                  & \multicolumn{2}{c}{Adam}                                                               \\
Batch Size                                                                 & \multicolumn{2}{c}{512}                                                                \\
Epochs                                                                     & \multicolumn{2}{c}{10}                                                                 \\
Learning Rate                                                              & \multicolumn{2}{c}{0.01}                                                               \\
Clipping Method                                                            & \multicolumn{2}{c}{Adaptive Gradient Clipping (AGC)}                                   \\
AGC Factor                                                                 & \multicolumn{2}{c}{0.01}                                                               \\ \hline
\end{tabular}
\end{table}

\begin{table}[h]
\caption{Parameters used to train the policy in SAC, SAC+FBO and \textit{SimKnoT}.}
\label{tab:models-sac}
\centering
\begin{tabular}{cc}
\hline
Parameter           & Value              \\ \hline
Initial Replay      & 100                \\
Batch Size          & 64                 \\
No. of Features     & 64                 \\
Warm up Transitions  & 100                \\
$\tau$                 & 0.005              \\
Learning Rate       & 0.001              \\
Learning Rate $\alpha$ & 0.001              \\
Discount Factor     & 0.99               \\
Optimizer           & Adam               \\
Critic Loss         & Mean Squared Error \\ \hline
\end{tabular}
\end{table}

\end{document}